\documentclass{article}

\usepackage{microtype}
\usepackage{graphicx}
\usepackage{subfigure}
\usepackage{booktabs} 

\usepackage{hyperref}

\usepackage[accepted]{icml2020}

\usepackage{amsmath}
\usepackage{amsfonts}
\usepackage{amsthm}
\usepackage{enumitem}
  \newenvironment{enumproof}
      { \begin{enumerate}[label={\textbf{\arabic*)}}, wide, labelwidth=!, labelindent=0pt] }
      { \end{enumerate} }

\newcommand{\mc}{\mathcal}
\newcommand{\attr}{{\rm at}}
\newcommand{\R}{\mathbb{R}}
\newcommand{\EE}{\mathbb{E}}
\newcommand{\prob}{\mathbb{P}}
\newcommand{\var}{{\rm Var}}

\newcommand{\norm}[1]{\left\lvert{#1}\right\rvert}
\newcommand{\dzvector}[1]{\boldsymbol{\mathbf{#1}}}
\renewcommand{\vec}[1]{\dzvector{#1}}

\newcommand{\supp}{{\rm supp}}
\newcommand{\bell}{{\rm Bell}}
\newcommand{\id}{{\rm id}}
\newcommand{\map}{{\vec z}}
\newcommand{\CEfour}{C_{\mc G}}

\newtheorem{theorem}{Theorem}
\newtheorem{proposition}{Proposition}
\newtheorem{lemma}{Lemma}
\newtheorem{define}{Definition}

\usepackage{xspace}
\newcommand{\appAffine}{\ref{sec:def-affine-inv-equiv}\xspace}
\newcommand{\appProofs}{\ref{sec:app:proofs}\xspace}
\newcommand{\appwGRNF}{\ref{sec:wGRNF}\xspace}
\newcommand{\appComplexity}{\ref{sec:comp-complexity}\xspace}

\begin{document}

\twocolumn[
\icmltitlerunning{Graph Random Neural Features for Distance-Preserving Graph Representations}
\icmltitle{Graph Random Neural Features\\ for Distance-Preserving Graph Representations}




\begin{icmlauthorlist}
\icmlauthor{Daniele Zambon}{usi}
\icmlauthor{Cesare Alippi}{usi,poli}
\icmlauthor{Lorenzo Livi}{win,ex}
\end{icmlauthorlist}

\icmlaffiliation{usi}{Universit\'{a} della Svizzera italiana, Lugano, Switzerland}
\icmlaffiliation{poli}{Politecnico di Milano, Milano, Italy}
\icmlaffiliation{win}{University of Manitoba, Winnipeg, Canada}
\icmlaffiliation{ex}{University of Exeter, Exeter, United Kingdom}

\icmlcorrespondingauthor{Daniele Zambon}{daniele.zambon@usi.ch}

\icmlkeywords{Machine Learning, ICML}

\vskip 0.3in
]



\printAffiliationsAndNotice{}  

\begin{abstract}
We present Graph Random Neural Features (GRNF), a novel embedding method from graph-structured data to real vectors based on a family of graph neural networks. 
The embedding naturally deals with graph isomorphism and preserves the metric structure of the graph domain, in probability.
In addition to being an explicit embedding method, it also allows us to efficiently and effectively approximate graph metric distances (as well as complete kernel functions); a criterion to select the embedding dimension trading off the approximation accuracy with the computational cost is also provided.
GRNF can be used within traditional processing methods or as a training-free input layer of a graph neural network.
The theoretical guarantees that accompany GRNF ensure that the considered graph distance is metric, hence allowing to distinguish any pair of non-isomorphic graphs. 
\end{abstract}

\section{Introduction}
Inference on graph-structured data is one of the hottest topics in machine learning, thanks to successes achieved in several scientific fields, like neurosciences, chemistry, computational biology and social sciences \cite{elton2019deep, battaglia2018relational, li2017fundamental}.
One of the major research challenges there consists of building a practical solution able to process graphs, yet managing the graph isomorphism problem. 
A way to address this latter problem passes through metric distances and complete kernels; however, it has been shown to be at least as hard as deciding whether two graphs are isomorphic~or not \cite{gartner2003graph}.

When data come as real vectors, the seminal paper by \citet{rahimi2008random} provides an efficient method to approximate radial basis kernels, exposing a parameter ---the embedding dimension--- trading off
computational complexity with approximation accuracy. 
The Random Kitchen Sinks \cite{rahimi2009weighted} technique builds on \cite{rahimi2008random} by adding a linear trainable layer and achieves, via convex optimization, an optimal estimation error on regression tasks \cite{rahimi2008uniform}; see also \cite{principe2015universal, rudi2017generalization} for discussions.
More recently, significant efforts aim at moving the research to the graph domain \cite{oneto2017measuring}, 
with most contributions focusing on graph neural network properties, especially on their ability to discriminate between non-isomorphic graphs \cite{chen2019equivalence,maron2019provably,xu2018powerful}.
Recent research also provided neural architectures granting the universal approximation property for functions on the graph domain \cite{maron2019universality,keriven2019universal}; the property holds asymptotically with the number of neurons.

\begin{figure*} 
\centering
\includegraphics[width=.95\textwidth]{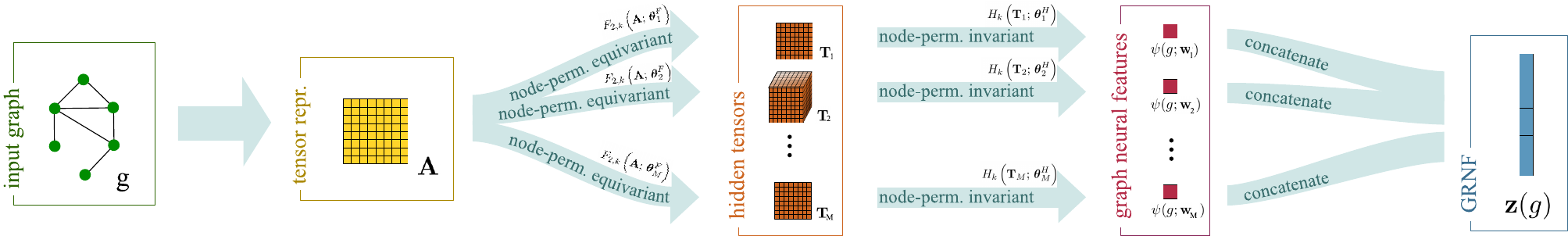}
\caption{Scheme of a GRNF map $\map:\mc G\rightarrow\R^M$. An $n$-node graph $g\in\mc G$ is firstly represented as a weighted adjacency matrix $\vec A\in\mc T^2$ (Section~\ref{sec:notation}). A collection of $M$ graph neural features $\{\psi(g;\vec w_m)\}_{m=1}^M$ is then computed, weighted and, finally, concatenated in vector $\map(g)$.
As described in Section~\ref{sec:gnf}, each graph neural feature map $\psi(\,\cdot\,;\vec w_m)$ is the composition of a node-permutation equivariant function, that maps matrix $\vec A$ to a tensor $\vec T_m\in\mc T^k $ of (potentially different) order $k$, and a node-permutation invariant one, that maps $\vec T_m$ to a scalar value $\psi(g;\vec w_m)\in\R$.}
\label{fig:graph-neural-features}
\end{figure*}

Here, we propose Graph Random Neural Features (GRNF), a training-free embedding method that provides a map $\map :\mc G\rightarrow\R^M$ from attributed graphs to numeric vectors and manages the graph isomorphism problem.
We prove that GRNF is able to discriminate between any pair of non-isomorphic graphs in probability and approximately preserves the metric structure of the graph domain in the embedding space. 
Notably, GRNF can also be employed as the first layer of a graph neural network architecture.
The main idea is to construct the map $\map$ from a family $\mc F=\{\psi(\,\cdot\,;\vec w)\ |\ \vec w \in\mc W\}$ of \emph{graph neural feature} maps $\psi(\,\cdot\,;\vec w):\mc G\rightarrow \R$, which are node-permutation-invariant graph neural networks with a single hidden-layer and a scalar output \cite{maron2018invariant}, separating%
\footnote{Family $\mc F$ is said to separate graphs of $\mc G$ when for any pair of distinct (non-isomorphic) graphs $g_1, g_2\in\mc G$, there exists a parameter vector $\vec w \in\mc W$
so that $\psi(g_1;\vec w)\neq \psi(g_2;\vec w)$.}
graphs in $\mc G$. 
Parameter vector $\vec w\in\mc W$ is randomly sampled from a suitable distribution $P$ and encodes the parameters associated with each neuron; as such, no training is requested.
Figure~\ref{fig:graph-neural-features} provides a visual description.

The major --novel-- theoretical contributions provided here are associated with Theorems \ref{theo:dP-metric} and \ref{theo:M-to-bounds}. Briefly,
\begin{itemize}
    \item 
    Theorem \ref{theo:dP-metric} shows that, given a suitable distribution $P$ over parameter set $\mc W$, distance
    \begin{equation}
    \label{eq:dP}
    d_P(g_1,g_2)=
        \left(
            \EE_{\vec w\sim P}
            \left[\left(\psi(g_1;\vec w) - \psi(g_2;\vec w)\right)^2\right]
        \right)^\frac{1}{2},
    \end{equation}
    is metric. This implies that $d_P(g_1,g_2)>0$ if and only if the two graphs are non-isomorphic.
    \item 
    Theorem \ref{theo:M-to-bounds} proves that the squared distance $\norm{\map(g_1)-\map(g_2)}_2^2$ between graph representations converges, by increasing the  embedding
dimension $M$, to $d_P(g_1,g_2)^2$ as $O(M^{-\frac{1}{2}})$, 
    and constructively suggests an embedding dimension $M_{\epsilon,\delta}$ that guarantees the discrepancy between the two to be less than an arbitrary value $\varepsilon$ with probability at least $1-\delta$. This guides the designer in selecting the embedding dimension $M$.
\end{itemize}
The paper is organized as follows. Section~\ref{sec:gnf} introduces the family $\mc F$ of graph neural features $\psi(\,\cdot\,;\vec w)$. Section~\ref{sec:g-dist} defines distance $d_P$ and proves Theorem~\ref{theo:dP-metric}. The GRNF map $\map$ is presented in Section~\ref{sec:GRNF}, where Theorem~\ref{theo:M-to-bounds} is also proven. Section~\ref{sec:related-work} relates our work with existing literature.
Finally, experimental results in Section~\ref{sec:experiments} empirically validate our theoretical developments.

\section{Notation}
\label{sec:notation}
For $k\in\mathbb{N}$, denote the space of order-$k$ tensors with size $n$ on every mode as
$$
\mc T^k = \R^{\overbrace{n\times \dots n\times n}^{k \text{ times}}};
$$
and $\R$ with $\mc T^0$.
Let us denote with $\pi\star\vec T$ the operation of applying permutation $\pi$, with $\pi$ in the symmetric group $S_n$, to each mode of tensor $\vec T\in\mc T^k$, namely $(\pi\star\vec T)_{i_1,\dots,i_k} = \vec T_{\pi(i_1),\dots,\pi(i_k)}$. In this paper we use the convention to represent with $\vec A$ tensors of order $2$, i.e., matrices, and with $\vec T$ tensors of generic order $k\in\mathbb{N}_0$. 

Let us define a graph $g$ with at most $n$ nodes as a triple $(V_g,E_g,\attr_g)$, with $V_g\subseteq \{1,2,\dots,n\}$ a set of nodes, $E_g\subset V_g\times V_g$ a set of edges, and attribute map $\attr_g:V_g\cup E_g\rightarrow \R$ associating nodes and edges with scalar attributes in a bounded subset of $\R$. 
We denote the set of such graphs with $\mc G(n)$, and with $\mc G$ the space $\bigcup_{n\in\mathbb{N}}\mc G(n)$ of graphs with arbitrarily large, yet finite order.
When no self-loops are present, we can represent each graph $g\in\mc G(n)$ with a tensor $\vec A_g \in\mc T^2$, where $(\vec A_g)_{i,i} = \attr_g(i)$, for $i\in V_g$, $(\vec A_g)_{i,j} = \attr_g((i,j))$, for $(i,j) \in E_g$ and $(\vec A_g)_{i,j} = 0$, otherwise. 
Different ordering choice of the nodes in $V_g$ results in different representations $\vec A_g$.
In our case, in which nodes are non-identified, this results in a bijection between $\mc G(n)$ and the quotient space ${\mc T^2}_{/S_n}=\{[\vec A]_{/S_n}\;|\;\vec A\in\mc T^k\}$ of equivalence classes $[\vec A]_{/S_n}=\{\pi\star \vec A\;|\; \pi\in S_n\}$.

Please note that the above assumptions are made to simplify the maths, however, extensions to allow for self-loops and attributes of any dimension is straightforward and detailed in Appendix~\appAffine.

\section{Graph neural features}
\label{sec:gnf}

\begin{figure*}
\centering
\includegraphics[width=.9\textwidth]{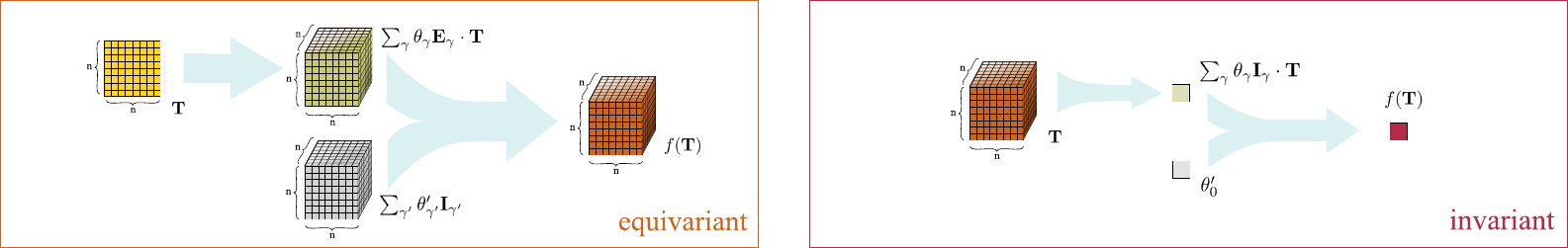}
\caption{Structure of an affine node-permutation equivariant function $F_{2,K}(\vec T,\vec \theta)$ (left) and a node-permutation invariant one $H_k(\vec T,\vec \theta)$ (right) as linear combinations of the bases $\{\vec I_\gamma\}$ and $\{\vec E_\gamma\}$ defined in Equation~\ref{eq:affine-inv-equiv}. In this example, $k=3$.}
\label{fig:affine-inv-equiv}
\end{figure*}

Given an integer $n\in\mathbb{N}$, we call \emph{graph feature map} any function $f:\mc T^2\rightarrow \R$ to the real set which is \emph{invariant} under permutation of the nodes, namely, $f(\vec A_g)=f(\pi\star \vec A_g)$, for every $\pi\in S_n$; indeed, having $g$ multiple representations $\vec A_g\in[\vec A_g]_{/S_n}$, this property is necessary to make $f(\vec A_g)$ a proper function of the graph $g$ itself, hence resulting with the same output regardless of the specific representation $\vec A_g$.
For this reason, in the following, we use the notation $f(g)$ and $f(\vec A_g)$ interchangeably.
Conversely, a function $f:\mc T^l\rightarrow \mc T^k$ is said \emph{equivariant} to node permutation if $f(\pi\star \vec T)=\pi\star f(\vec T)$, $\forall\pi\in S_n$.

Recent findings \cite{maron2018invariant} have shown that the set of all linear permutation-invariant functions $\mc T^k\rightarrow \R$ is a vector space of dimension $\bell(k)$, i.e., the number of partitions of set $\{1,2,\dots, k\}$; similarly, the space of linear permutation-equivariant functions is proven to be a $\bell(k+l)$-dimensional vector space. Denoting with $\{\vec I_\gamma\}_{\gamma=1}^{\bell(k)}$ and $\{\vec E_\gamma\}_{\gamma=1}^{\bell(k+l)}$ the bases of invariant and equivariant linear spaces, respectively, we obtain that every affine invariant and equivariant function can be written in terms of tensor products and sums of the form 
\begin{equation}
\label{eq:affine-inv-equiv}
f(\vec T) =
\begin{cases}
H_k(\vec T; \vec \theta) = \sum_{\gamma} \theta_\gamma \vec I_\gamma \vec T + \theta'_0 \\
F_{l,k}(\vec T; \vec \theta) = \sum_{\gamma} \theta_\gamma   \vec E_\gamma \vec T + \sum_{\gamma'} \theta'_{\gamma'} \vec I_{\gamma'}
\end{cases}
\end{equation}
where $\{\theta_\gamma\}$ and $\{\theta'_{\gamma'}\}$ are coefficients that relate to the linear and bias parts, respectively, and $\vec \theta = \{\theta_\gamma\}\cup\{\theta'_{\gamma'}\}$.
We refer the reader to Appendix~\appAffine for a proper definition of $\{\vec I_\gamma\}$ and $\{\vec E_\gamma\}$; Figure~\ref{fig:affine-inv-equiv} provides a visual representation of these affine functions.

We are ready to define the set of graph neural features as composition of equivariant and invariant affine maps and some (nonlinear) activation functions.
\begin{define}[Graph neural feature]
\label{def:graph-neural-feature}
We define a \emph{graph neural feature} to be a parametric map $\psi(\;\cdot\;;\vec w):\mc T^2\rightarrow\R$, with parameters $\vec w=(k,\vec \theta_F,\vec \theta_H)$, and resulting from the composition 
$$
\mc T^2 
\xrightarrow{F_{2,k}(\;\cdot\;; \vec \theta_F)}
\mc T^k
\xrightarrow{\rho_e}
\mc T^k
\xrightarrow{H_{k}(\;\cdot\;; \vec \theta_H)}
\R
\xrightarrow{\rho_i}
\R,
$$
where $\rho_i$, $\rho_e$ are activation functions applied component wise and
$F_{2,k}(\cdot;\vec\theta_F)$, $H_{k}(\cdot;\vec \theta_H)$ are affine equivariant and invariant ones, respectively, in the form of Equation~\ref{eq:affine-inv-equiv}. 
The parameter space is
\begin{equation}
    \label{eq:parameter-space}
\mc W = \left\{\vec w=(k,\vec \theta_F,\vec \theta_H)\right\},
\end{equation}
for $k\in\mathbb{N}, \ \vec \theta_F\in\R^{\bell(k+2)+\bell(k)},\ \vec \theta_H\in\R^{\bell(k)+1}$.
\end{define}
Note that by composing component-wise any activation function $\rho_i$, $\rho_e$ --e.g., the sigmoid-- to an affine invariant (equivariant) function gives an invariant (equivariant) function. Therefore, in the end $\psi(\,\cdot\,;\vec w)$ is an invariant function. Without any ambiguity, we can write $\psi(g;\vec w)$ as function of $g$.
We comment that, despite $\psi$ has been introduced for graphs in $\mc G(n)$ with at most $n$ nodes, it is readily extended to operate on the entire space $\mc G$ of graphs with arbitrarily large order, in fact, parameter space $\mc W$ does not change with respect to $n$ \cite{maron2018invariant}; this is crucial to compare graphs of different orders.
We denote with 
\begin{equation}
    \label{eq:def-graph-neural-feature-map-space}
    \mc F(\rho_i,\rho_e, \mc W),
\end{equation}
or simply $\mc F$, the set of all graph neural feature maps introduced above and defined over the entire $\mc G$. 
A generalization to graph with vector attributes associated to nodes and edges is possible, as well (please, see to Appendix~\appAffine).

As shown in the following Lemma~\ref{lemma:F1-separates}, family $\mc F$ is rich enough to separate graphs of $\mc G$; in other terms, its expressive power permits
to distinguish any pair of non-isomorphic graphs. 
\begin{lemma}
\label{lemma:F1-separates}
Let $\rho_e$ be a squashing function, and $\rho_i$ a non-constant one.
Then set $\mc F(\rho_i,\rho_e,\mc W)$ is rich enough to separate graphs of $\mc G$, namely, for any pair of distinct (non-isomorphic) graphs $g_1, g_2\in\mc G$, there exists a parameter configuration $\vec w^*\in\mc W$ so that $\psi(g_1;\vec w^*)\neq \psi(g_2;\vec w^*)$.
\end{lemma}
Note that the sigmoid function $\sigma(x)=(1+e^{-x})^{-1}$ is both squashing and non-constant, therefore family $\mc F(\sigma, \sigma, \mc W)$ of graph neural features
$$\psi(\cdot,\vec w) = \sigma\circ H_k(\cdot,\vec \theta_H)\circ \sigma \circ F_{2,k}(\cdot, \vec \theta_F) $$
fulfills the hypothesis of Lemma~\ref{lemma:F1-separates}.

The proof proceeds with the same strategy adopted by \citet{hornik1989multilayer} to exhibit approximation capabilities of multi-layer perceptrons and, recently, by \citet{keriven2019universal} to graph neural networks; see Appendix~\appProofs for details.

\section{A metric distance for graphs}
\label{sec:g-dist}

The family $\mc F$ of graph neural features \eqref{eq:def-graph-neural-feature-map-space} allows to define the distance $d_P(g_1,g_2)$ of Equation~\ref{eq:dP}. The distance assesses the expected discrepancy between graph neural features associated with two graphs
$g_1, g_2\in\mc G$; it is only requested that $\EE_{\vec w}[\psi(g;\vec w)^2]<\infty$ for every $g\in\mc G$.
Notice that the distance depends on the distribution $P$. As such, a change in $P$ results in a different distance. In a supervised setting, this is a ``parameter'' that can be tuned (see subsequent Section~\ref{sec:GRNF}).
One of the simplest examples of distance $d_P$ originates from considering a uniform distribution over a finite set $\{\tilde{\vec w}_r\}_{r=1}^R$ of $R\in\mathbb{N}$ parameters. This specific choice gives 
$$
    d_P(g_1,g_2)=\left(\frac{1}{R}\sum_r \left(\psi(g_1;\tilde{\vec w}_r)-\psi(g_2;\tilde{\vec w}_r)\right)^2\right)^\frac{1}{2},
$$ 
and results in a pseudo-metric distance, as it is positive and symmetric, satisfies the triangular inequality. However, the \emph{identifiability} property 
\begin{equation}
\label{eq:identifiability}
  g_1 = g_2 \iff d_{P}(g_1, g_2)=0, \qquad \forall\ g_1,g_2\in\mc G
\end{equation} 
does not hold.
It can be proved%
    \footnote{Please, see proof of Theorem~\ref{theo:dP-metric}.}
that the resulting distance $d_P$ is always at least a pseudo-metric, regardless of the choice of $P$.

Remarkably, a principled choice of $P$, such that its support $\supp(P)$ is the entire set $\mc W$, ensures that $d_P$ is metric, as Theorem~\ref{theo:dP-metric} guarantees. 
\begin{theorem}
\label{theo:dP-metric}
Consider set $\mc F(\rho_i,\rho_e, \mc W)$ of graph neural features on the graph set $\mc G$.
Define a probability distribution $P$ on $\mc W$, and the corresponding graph distance $d_P$ according to \eqref{eq:dP}.
Under the following assumptions, space $(\mc G, d_{P})$ is metric.
\begin{enumerate}[label=(\textbf{A\arabic*})]
    \item \label{ass:rho} 
    Functions $\rho_i,\rho_e:\R\rightarrow\R$ are continuous, with $\rho_e$ being a squashing function and $\rho_i$ a non-constant one;
    \item \label{ass:suppP} 
    Support $\supp(P)$ of $P$ covers $\mc W$;
    \item \label{ass:forth} 
    There exists a positive constant $\CEfour$ such that the fourth momentum $\EE_{\vec w}[\psi(g;\vec w)^4]<\CEfour$, $\forall$ $g\in\mc G$. 
\end{enumerate} 
\end{theorem}

Notice that all these assumptions are ``only'' sufficient conditions, and are rather mild. In fact, one practical choice to satisfy them all at once is to build distribution $P$ over a Poisson distribution to select the tensor order $k$ and a corresponding multivariate Gaussian distribution for parameter vectors $\vec \theta_H,\vec \theta_F$, and consider the sigmoid function for both $\rho_i$ and $\rho_e$.
Moreover, without loss on generality, we can assume $\CEfour=1$. 

The core of the proof aims at showing that $d_P$ possesses the identifiability property \eqref{eq:identifiability}.
To sketch the proof, notice that Assumption \ref{ass:rho} enables Lemma~\ref{lemma:F1-separates} and ensures that for any pair of graphs $g_1\ne g_2$ there exists a particular $\tilde{\vec w}\in\mc W$ for which $\psi(g_1;\tilde{\vec w})\neq \psi(g_2;\tilde{\vec w})$.
On the other hand, Assumption \ref{ass:suppP} grants that every feature map in $\mc F$ is taken into account by \eqref{eq:dP}, and together with \ref{ass:rho}, that we can find a neighbourhood $U(\tilde{\vec w})$ of non-null probability for which
$$
  \psi(g_1;\vec w)\neq \psi(g_2;\vec w), \qquad \forall\vec w\in U(\tilde{\vec w}).
$$
Property~\eqref{eq:identifiability} follows from $(\psi(g_1,\cdot) - \psi(g_2,\cdot))^2\ge 0$.

Finally, showing that $d_P$ is always positive, symmetric, and satisfies the triangular inequality is easier. Observe in fact that for every random variable $X_1,X_2$, we have 
$\EE[(X_1+X_2)^2]\leq (\sqrt{\EE[X_1^2]}+\sqrt{\EE[X_2^2]})^2$, and in particular for $X_1=\psi(g_1;\vec w)-\psi(g_3;\vec w)$ and $X_2=\psi(g_3;\vec w)-\psi(g_2;\vec w)$.
A detailed proof is provided in Appendix~\appProofs.

\section{A complete kernel for graphs}
\label{sec:g-kernel}

The family $\mc F$ of graph neural feature maps in \eqref{eq:def-graph-neural-feature-map-space} allows to define also a kernel function for graphs, hence to process graph in the well-grounded theory of reproducing kernel Hilbert spaces.

Following the same rationale used for distance $d_P$ in Section~\ref{sec:g-dist}, we define the following positive-definite kernel 
\begin{equation}
\label{eq:kP}
    \kappa_{P}(g_1,g_2) := \EE_{\vec w\sim P}\left[ \xi(g_1;\vec w)\, \xi(g_2;\vec w)\right],
\end{equation}
where $\xi(g_1,\vec w) = \psi(g_1;\vec w) - \psi(\vec 0_g;\vec w)$ and $\vec 0_g$ is a {``null''} graph, which can be represented by the adjacency $\vec A=0\in\R^{1\times 1}$ of a graph with a single node. Kernel $\kappa_P$ is intimately related with $d_P$, in fact:
\begin{equation}
\label{eq:d-kkk}
d_P(g_1,g_2)^2 = \kappa_P(g_1,g_1)-2\kappa_P(g_1,g_2)+\kappa_P(g_2,g_2), 
\end{equation}
or, alternatively, we can express $k_P$ as induced by $(d_P)^2$
\begin{equation}
\label{eq:k-ddd}
\kappa_P(g_1,g_2) = \frac{1}{2}\left(d_P(g_1,\vec 0)^2 +d_P(g_2,\vec 0)^2 - d_P(g_1,g_2)^2\right).
\end{equation}
Proposition~\ref{prop:neg-distance-complete-kernel} shows that $\kappa_P$ in \eqref{eq:kP} is a \emph{complete} kernel, meaning that the canonical embedding $\phi:\mc G\rightarrow \mc H$ to the reproducing kernel Hilbert space $\mc H$, and for which we can write $\kappa(g_1,g_2)=\langle \phi(g_1),\phi(g_2)\rangle_{\mc H}$, is injective \cite{gartner2003graph}, thus it maps distinct graphs to different points in $\mc H$. 
\begin{proposition}
\label{prop:neg-distance-complete-kernel}
Under Assumptions~\ref{ass:rho}--\ref{ass:forth}, we have:
\begin{itemize} 
\item 
$d_P$ is of \emph{negative type}, i.e., for every $S\in\mathbb{N}$, any set of graphs $\{g_s\}_{s=1}^S$ and any set of scalars $\{c_s\}_{s=1}^S$, with $\sum_s c_s = 0$, it follows
$\sum_{i}\sum_{j} c_i\,c_j\;d_P(g_i,g_j) \le 0;$
\item
$\kappa_P$ is a \emph{complete} graph kernel.
\end{itemize}
\end{proposition}

The proof follows from the work of \citet{sejdinovic2013equivalence}, where if $\kappa_P$ can be written in the form of \eqref{eq:k-ddd} 
and $(d_P)^2$ is a semi-metric of negative type, then (i) $d_P$ is of negative type, as well [Prop. 3], and (ii) $\kappa_P$ is a complete kernel [Prop. 14].

We now need to show that the semi-metric $(d_P)^2$ is of negative type.
Please, notice that kernel $\kappa_P$ is positive semi-definite, in fact, for any choice of $S,\{c_s\},\{g_s\}$
\begin{equation}
\label{app:eq:lemma:kP-is-pd}
\sum_{i,j=1}^S c_ic_j\,k_{P}(g_i,g_j) 
    =\EE_{\vec w\sim P}\left[f(\vec w)^2\right]\geq 0.
\end{equation}
where $f(\vec w)=\sum_{s=1}^S c_s\xi(g_s;\vec w)$.
From relation \eqref{eq:d-kkk}, and when $\sum_s c_s = 0$,
\begin{align*}
\sum_{i,j=1}^S c_i c_j d_P(g_i,g_j)^2  
&= 0-2\sum_{i,j=1}^S c_i c_j \kappa_P(g_i,g_j) + 0 \le 0,
\end{align*}
thanks to \eqref{app:eq:lemma:kP-is-pd}. This concludes the thesis.

\section{Graph random neural features}
\label{sec:GRNF}

The definition of graph distance $d_P$ in \eqref{eq:dP} as an expectation over the feature maps in $\mc F$ allows us to create a random mapping that approximately preserves the metric structure of graph space $(\mc G,d_P)$.
\begin{define}[Graph Random Neural Features (GRNF)]  
Given probability distribution $P$ defined over $\mc W$ and an embedding dimension $M\in\mathbb{N}$, we define \emph{Graph Random Neural Features map} a function $\map:\mc G\rightarrow \R^M$ that associates to graph $g\in\mc G$ the vector
\begin{equation}
\label{eq:GRNF-plain}
\map(g; \vec W):=\frac{1}{\sqrt{M}}\left[\psi(g;\vec w_1),\dots,\psi(g;\vec w_M)\right]^\top,
\end{equation}
where $\vec W=\{\vec w_m\}_{m=1}^M$ are drawn independently from $P$, and $\{\psi(\cdot;\vec w_m)\}_{m=1}^M\subseteq\mc F(\rho_i,\rho_e,\mc W)$ are graph neural features (Definition~\ref{def:graph-neural-feature}).
\end{define}
In the following, we may omit the explicit dependence from $\vec W$ in $\map(g; \vec W)$, when clear from the context, and use the compact notation $\map(g)$.
The computational complexity of \eqref{eq:GRNF-plain} is studied in Appendix~\appComplexity.
Figure~\ref{fig:graph-neural-features} provides a visual representation of the map.

Along the same lines of random Fourier features~\cite{rahimi2008random,li2019towards} ---but focusing on a distance rather than on a kernel--- we have that the squared norm $\norm{\map(g_1)-\map(g_2)}_2^2$, which can be thought as a sample mean of different graph neural features, is an unbiased estimator of the squared distance $d_P(g_1,g_2)^2$. Moreover, thanks to the law of large numbers, we have
\begin{equation}
\label{eq:approx-distance}
    \norm{\map(g_1)-\map(g_2)}_2^2 \stackrel{p}{\rightarrow} d_P(g_1,g_2)^2,
\end{equation} 
where the convergence is in probability and as $M\rightarrow\infty$, i.e., for any $\varepsilon>0$
$$\lim_{M\rightarrow \infty}\prob\left(\norm{\norm{\map(g_1)-\map(g_2)}_2^2 - d_P(g_1,g_2)^2}>\varepsilon\right) =0.$$
By continuity, we also have $\norm{\map(g_1)-\map(g_2)}_2 \stackrel{p}{\rightarrow} d_P(g_1,g_2)$. 
Since we proved that distance $d_P$ is metric, the above convergence shows the ability of GRNF to distinguish all non-isomorphic graphs.
The convergence \eqref{eq:approx-distance} is of order $\sqrt{M}$, as one can see by 
\begin{multline}
\var\left[\norm{\map(g_1)-\map(g_2)}_2^2\right] = \\
=\frac{1}{M^2}\sum_{m=1}^M \var\left[\left(\psi(g_1;\vec w)-\psi(g_2;\vec w)\right)^2\right]
\\\label{eq:bound-variance}
  \le \frac{8}{M} \left( \EE\left[\psi(g_1;\vec w)^4\right] + \EE\left[\psi(g_2;\vec w)^4\right]\right)
\stackrel{\ref{ass:forth}}{\le} \frac{16}{M},
\end{multline}
thanks%
  \footnote{Being $x^p$ convex, we have $\left(\frac{X}{2}+\frac{Y}{2}\right)^p \leq \frac{X^p}{2} +\frac{Y^p}{2}$, therefore $\frac{1}{2^p}\EE\left[\left(X+Y\right)^p\right] \leq \frac{1}{2} (\EE[X^p] +\EE[Y^p])$.
  }
to the convexity of $g(x)=x^4$. Finally, by exploiting \eqref{eq:bound-variance} and Chebyshev's inequality, the following Theorem~\ref{theo:M-to-bounds} follows.
\begin{theorem}
\label{theo:M-to-bounds}
Under Assumption~\ref{ass:forth}, for any value $\varepsilon>0$ and $\delta\in(0,1)$, inequality
$$\prob\left(\norm{\norm{\map(g_1)-\map(g_2)}_{2}^2 -d_{P}(g_1,g_2)^2} \ge \varepsilon\right) \le \delta$$
holds with embedding dimension $M\geq \frac{16}{\delta\,\varepsilon^2}$.
\end{theorem}
Theorem~\ref{theo:M-to-bounds} allows to identify an embedding dimension $M$ that ensures to fulfill some application requirements expressed in terms of $\varepsilon$ and $\delta$.
An analogous result:
$$
\prob
    \left(
        \norm{
        \widetilde{\kappa}_{P}(g_1,g_2) 
        -  \kappa_{P}(g_1,g_2)} 
        \ge \varepsilon
    \right) \le \delta
$$
is proven also for the approximated kernel
$\widetilde{\kappa}_{P}(g_1,g_2)$ $=(\map(g_1)-\map(\vec 0_g))^\top(\map(g_2)-\map(\vec 0_g))$, provided that $M\ge \frac{1}{\delta\,\varepsilon^2}$.
Again, this is a consequence of convergence
\begin{equation}
\label{eq:approx-kernel}
    \widetilde{\kappa}_(g_1,g_2)\stackrel{p}{\rightarrow} \kappa_P(g_1,g_2),
\quad M\rightarrow\infty.
\end{equation} 

Remarkably, we can obtain the same convergence results also when sampling the parameters $\vec W=\{\vec w_1,\dots,\vec w_M\}$ from a distribution $\overline P$ different from $P$. It is only necessary to appropriately weight each component of map $\map$. In compliance with \cite{li2019towards}, we call \emph{Weighted GRNF}
\begin{equation}
\label{eq:GRNF-weighted}
\overline\map(\;\cdot\;;\vec W)
:=\left[\dots, \sqrt{\frac{p(\vec w_m)}{M\,\overline p(\vec w_k)}}\psi(\;\cdot\;;\vec w_m),\dots\right]^\top 
\end{equation}
where $p$ and $\overline p$ are the mixed-type probability distributions%
    \footnote{We consider that random variable $\vec w=(k,\vec \theta_H,\vec \theta_F)\sim P$ is composed a discrete component $k$ with probability mass function $p_k(k)$ and a continuous part $(\vec \theta_H,\vec \theta_F)$ with probability density function $p_\theta(\vec \theta_H,\vec \theta_F|k)$. Finally, we define with $p(\vec w)=p_k(k)p_\theta(p_\theta(\vec \theta_H,\vec \theta_F|k)$.}
associated with $P$ and $\overline P$, respectively.
We refer the reader to Appendix~\appwGRNF for further details on this interesting setting.

The specific choice of distribution $P$ induces different distances $d_P$ (and kernel functions $\kappa_P$), and a principled choice of $P$ can make the difference.
In practice, 
we can exploit the trick of sampling from a predefined distribution $\overline P$, as in \eqref{eq:GRNF-weighted}, and, a posteriori, identify a suitable distribution $P$ that best serves the task to be solved \cite{sinha2016learning}. Specifically, once $M$ parameter vectors $\{\vec w_m\}_{m=1}^M$ are sampled from $\overline P$, we have that in \eqref{eq:GRNF-weighted} the scalars $\{p_m=p(\vec w_m)\}_{m=1}^M$ become free parameters.

To conclude the section, we stress that GRNF can trade-off metric distances and complete kernels for practical computability, as described by Theorem \ref{theo:M-to-bounds}, in agreement with \citet{gartner2003graph}.

\section{Related work}
\label{sec:related-work}

The process of random sampling features takes inspiration from the work on random Fourier features~\cite{rahimi2008random}. There, shift-invariant kernels $\kappa(\vec x_1,\vec x_2)=\kappa(\vec x_1-\vec x_2)$ are written in the form of an expected value 
\begin{equation}
\label{eq:random-fourier-kernel}
    \EE_{\omega\sim p}[\zeta(\vec x_1;\vec\omega)\zeta(\vec x_2;\vec\omega)^*]
\end{equation}
with $p$ being the Fourier transform of $\kappa(\cdot)$ and $\zeta(\vec x;\vec\omega)=e^{i\vec \omega^\top \vec x}$ (Bochner's theorem \cite{rudin1991functional}).
By random sampling parameters $\{\vec \omega_m\}_{m=1}^M$ from $p$ and $\{b_m\}_{m=1}^M$ from the uniform distribution $U([0,2\pi))$, we have that 
\begin{multline*}
\EE\left[\frac{1}{M}\sum_m\cos(\vec \omega_m^\top \vec x_1 + b_m)\cos(\vec \omega_m^\top\vec  x_2 + b_m)\right]
\\ =\kappa(\vec x_1,\vec x_2),
\end{multline*}
hence providing a Monte Carlo approximation method for the kernel.
Clearly, being able to compute the Fourier transform $p$ is crucial, however, it is not always possible; see, e.g., \citet{vedaldi2012efficient} for some examples.
Extensions of this approach consider dot-product kernels $\kappa(\vec x_1,\vec x_2)=\kappa(\vec x_1^\top \vec x_2)$ \cite{kar2012random}.
The work by \citet{yu2016orthogonal} proposed to keep an orthogonal structure in the matrix of sampled weights in order to achieve better approximations with a smaller number of features.
Alternatives to random sampling employ Gaussian quadrature to approximates the kernel with higher a convergence rate \cite{dao2017gaussian}.
Finally, we mention the works by \citet{wu2018d2ke,wu2019scalable} which, relying on a distance measure between data points, can apply the same rationale also to non-numeric data. 

Our proposal builds on the framework of random features, but it works somehow in the opposite direction. Firstly, it defines a parametric family of features $\mc F=\{\psi(\,\cdot\,;\vec w)|\vec w\in\mc W\}$ that separates graphs of $\mc G$ ---playing the role of $\{\zeta(\,\cdot\,;\vec \omega)\}$ in \eqref{eq:random-fourier-kernel}--- and, subsequently, it provides a distance (and kernel) function by selecting a probability distribution $P$ on parameter space $\mc W$. 
This choice has two major advantages: (1) it does not require to compute distribution $P$ from a $\kappa$, and (2) allows the selection of the most appropriate distribution based on the task at hand. Moreover, the same rationale can be applied to other families of features that separate graphs.

\begin{figure}
    \centering
    \includegraphics[width=.48\columnwidth]{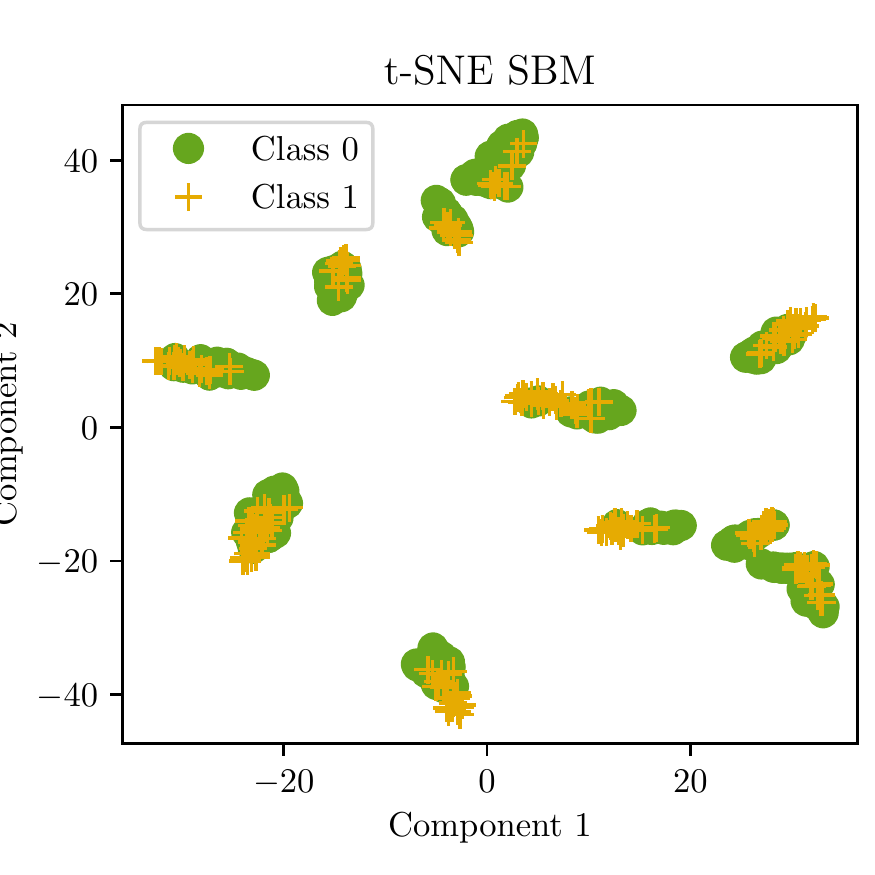}
    \includegraphics[width=.48\columnwidth]{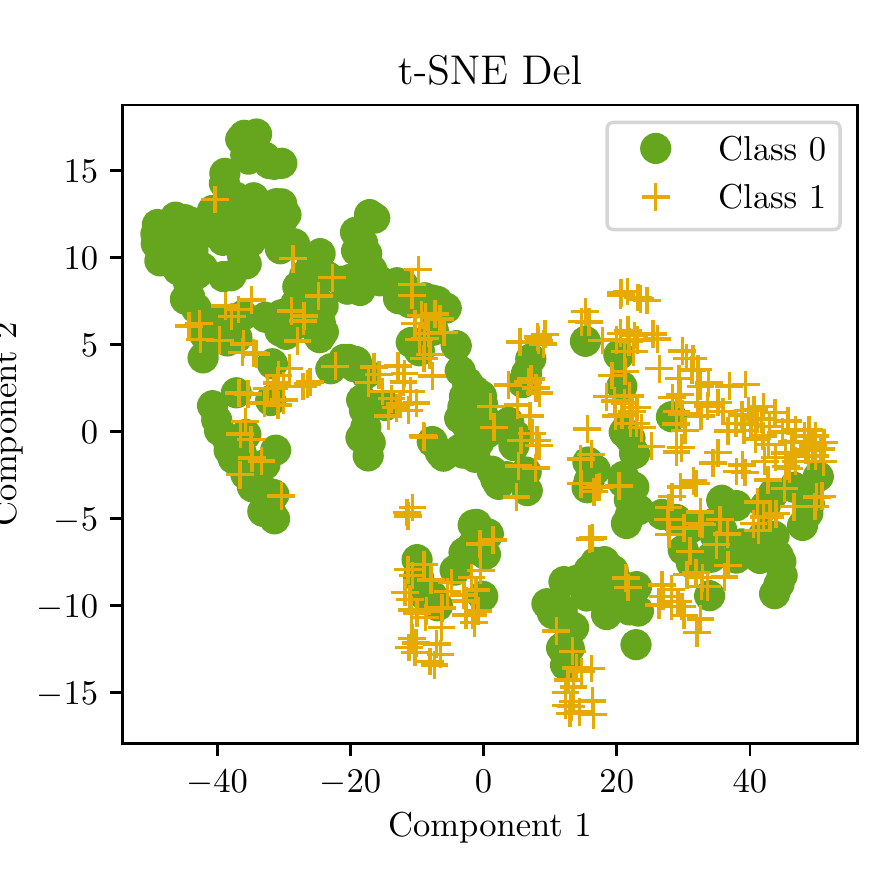}
    \caption{2-dimensional visualization of the embedding vectors $\vec z_*(g_i)$ for $i=1,\dots, 600$ drawn with t-SNE.}
    \label{fig:tsne}
\end{figure}

\begin{figure}
    \centering
    \includegraphics[width=.94\columnwidth]{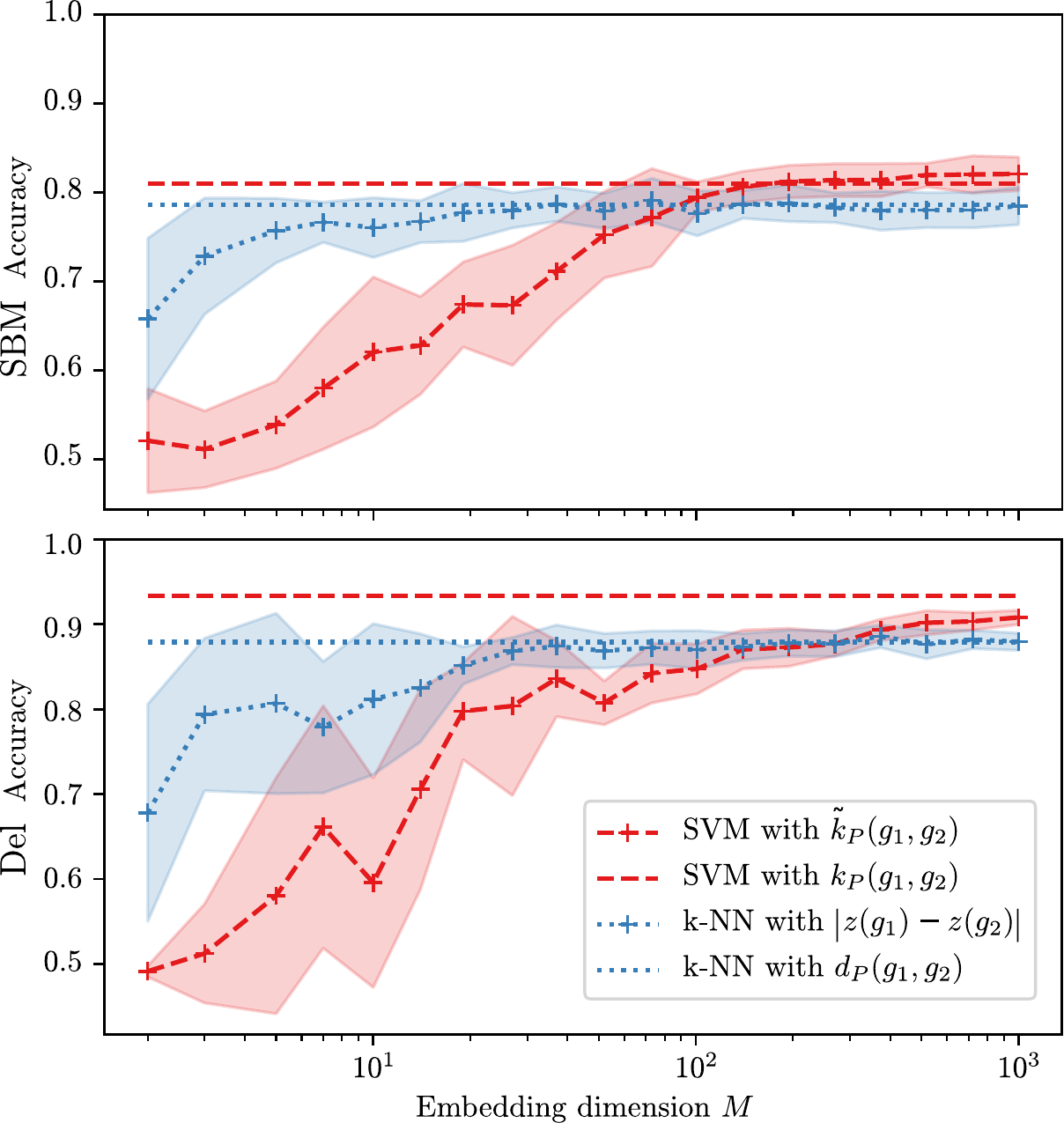}
    \caption{Classification performance in terms of accuracy using different embedding dimensions $M$. First and second rows correspond to data sets SBM and Del, respectively. Each reported accuracy value is an average across 10 repetitions and the shaded region represents one standard deviation from the average.}
    \label{fig:classif-del-sbm}
\end{figure}

\begin{table*}  
\centering
\caption{We report accuracy and standard deviation estimated on 10-fold cross-validation, where in each run we consider the optimal hyper-parameter configuration assessed on a validation set.
Results from Baseline, DGCNN, DiffPool, ECC, GIN and GraphSAGE are reported from \cite{errica2020fair}. (Notice that the authors created two versions of the social data sets, one with no node attributes and the other augmented with node degrees as node feature. Here, we considered the former set up).}
\label{tab:gnn-comparison}
\begin{tabular}{r|ccc|ccc}
\hline 

\hline 
 & \textbf{NCI1} & \textbf{PROTEINS} & \textbf{ENZYMES} & \textbf{IMDB-BINARY} & \textbf{IMDB-MULTI} & \textbf{COLLAB} \\
\hline 
Baseline  & 69.8$\pm$2.2 & 75.8$\pm$3.7 & 65.2$\pm$6.4 & 50.7$\pm$2.4 & 36.1$\pm$3.0 & 55.0$\pm$1.9 \\
\hline 
DGCNN     & 76.4$\pm$1.7 & 72.9$\pm$3.5 & 38.9$\pm$5.7 & 53.3$\pm$5.0 & 38.6$\pm$2.2 & 57.4$\pm$1.9 \\
DiffPool  & 76.9$\pm$1.9 & 73.7$\pm$3.5 & 59.5$\pm$5.6 & 68.3$\pm$6.1 & 45.1$\pm$3.2 & 67.7$\pm$1.9 \\
ECC       & 76.2$\pm$1.4 & 72.3$\pm$3.4 & 29.5$\pm$8.2 & 67.8$\pm$4.8 & 44.8$\pm$3.1 & --- \\
GIN       & 80.0$\pm$1.4 & 73.3$\pm$4.0 & 59.6$\pm$4.5 & 66.8$\pm$3.9 & 42.2$\pm$4.6 & 75.9$\pm$1.9 \\
GraphSAGE & 76.0$\pm$1.8 & 73.0$\pm$4.5 & 58.2$\pm$6.0 & 69.9$\pm$4.6 & 47.2$\pm$3.6 & 71.6$\pm$1.5 \\
\hline
GRNF      & 66.7$\pm$2.4 & 75.1$\pm$2.6 & 45.9$\pm$6.6 & 69.7$\pm$3.8 & 44.4$\pm$3.9 & 73.2$\pm$2.5 \\
\hline 

\hline 
\end{tabular}
\end{table*}

\begin{figure}
    \centering
    \includegraphics[width=.9\columnwidth]{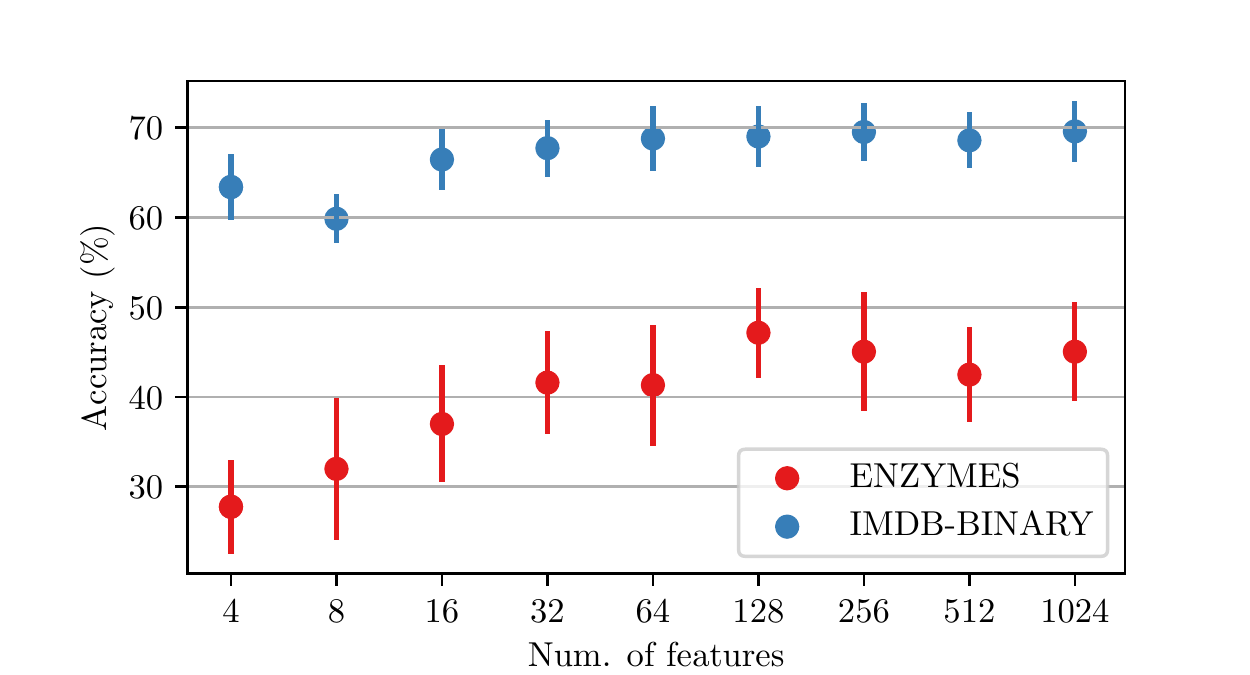}
    \caption{Classification performance on ENZYMES and IMDB-BINARY data sets in terms of accuracy using different embedding dimensions $M$. Each reported accuracy value is an average across 10 repetitions and the bar indicate one standard deviation from the average.}
    \label{fig:convergence-TU}
\end{figure}

\section{Experimental validation}
\label{sec:experiments}

The experimental campaign is divided into two parts. Section~\ref{sec:exp-convergence} gives empirical evidence about the claimed convergence as the embedding dimension $M$ grows. Secondly, Section~\ref{sec:exp-gnn} shows that our method can be effectively used as a layer of a neural network and achieves results comparable to the current state of the art on classification tasks. 

\subsection{Convergence increasing the embedding dimension}
\label{sec:exp-convergence}

We considered two data sets with two classes: one contains simple graphs with no attributes, and the other one has graphs with bi-dimensional attributes associated to each node. Figure~\ref{fig:tsne} shows a visualization by means of t-SNE \cite{maaten2008visualizing} to perceive the complexity of the classification problem.

\paragraph{Graphs from the stochastic block model (SBM).}
We generated two classes of 300, 12-node graphs from the stochastic block model \cite{holland1983stochastic}. Class 0 has a single community with edge probability 0.4, while class 1 has two communities of 6 nodes with 0.8 probability of connecting two nodes of the same community, while the probability of connecting nodes of different communities equals 0.1.

\paragraph{Delaunay's triangulation graphs (Del).}
We generated two classes of 300, 12-node Delaunay's triangulation graphs \cite{zambon2017detecting,grattarola2019change}.
The graphs of a single class have been generated starting from a collection of 6 planar points, called seed points (the two classes are determined by different collections of seed points). 
Seed points are then perturbed with Gaussian noise. Each point corresponds to a node of the graph and its coordinates are considered as node attributes. Finally, the Delaunay's triangulation of the perturbed points gives the topology of the graph. 

\begin{figure} 
\centering
\includegraphics[width=.9\columnwidth]{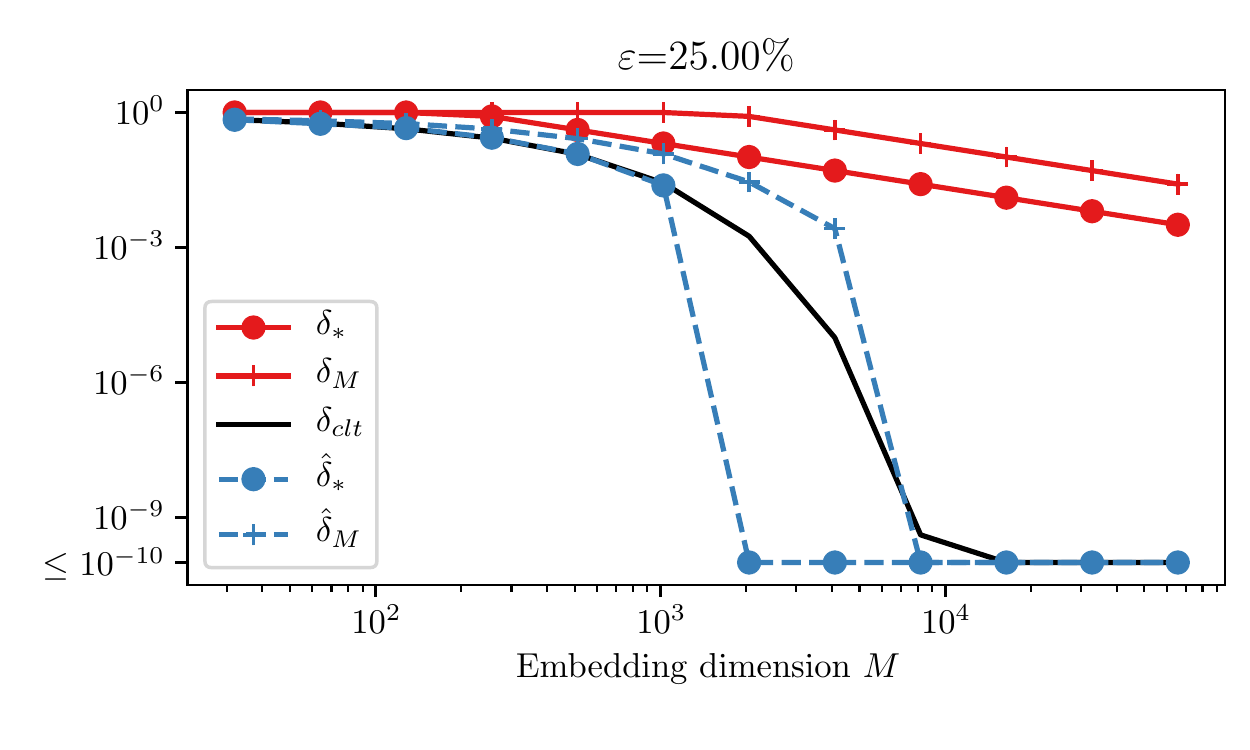}
\caption{Empirical verification of Theorem~\ref{theo:M-to-bounds}. $\delta_M$, $\delta_*$ and $\delta_{clt}$ defined as in Eq.~\ref{eq:delta_M}, \ref{eq:delta_star} and \ref{eq:delta_clt}, respectively. $\hat \delta_M$ and $\hat \delta_*$ denote the empirical computation of the left-hand side in Eq.~\ref{eq:delta_M} and \ref{eq:delta_star}, respectively. Value of $\varepsilon$ is set to the 25\% of the squared graph distance computed with $M_*$.}
\label{fig:verify-theo}
\end{figure}

The first experiment provides empirical evidence of the validity of the bounds in Theorem \ref{theo:M-to-bounds}.
Since it is not always possible to compute the true value of $d_P$, we make use of two GRNF, $\map(\,\cdot\,;\vec W_1)$ and $\map(\,\cdot\,;\vec W_2)$, both with embedding dimension $M$. Let be $\Delta(\vec W)=\norm{\map(g_1,\vec W)-\map(g_2,\vec W)}_2^2$, then%
\footnote{
  For any $\alpha\in(0,1)$, it holds true that $\prob(|A-B|\ge\varepsilon)\le \prob(|A-C|+|C-B|\ge\varepsilon)\le\prob(|A-C|\ge \varepsilon\,\alpha) + \prob(|C-B|\ge\varepsilon\,(1-\alpha))$. By substituting $\alpha=\frac{1}{2}$, $A=\Delta(\vec W_1)$, $B=\Delta(\vec W_2)$ and $C=d_P(g_1,g_2)^2$, we obtain \eqref{eq:delta_M}.
}
\begin{equation}
\label{eq:delta_M}
    \prob\left(\norm{\Delta(\vec W_1)-\Delta(\vec W_2)}\ge \varepsilon\right)
    \le  \frac{128}{M\,\varepsilon^2} =:\delta_M.
\end{equation}
We also compared the above $M$-dimensional approximation $\Delta$ with a better estimate $\Delta_*:= \norm{\map_*(g_1)-\map_*(g_2)}_2^2$ based on a $M_*$-dimensional map $\map_*$ with $M_*=10^6\gg M$. Assuming that equation $\Delta_*=d_P(g_1,g_2)^2$ holds
\begin{equation}
\label{eq:delta_star}
\prob\left(\norm{\Delta-\Delta_*}\ge \varepsilon\right)
\le \frac{16}{M\,\varepsilon^2} =:\delta_*.
\end{equation}
Finally, we have performed a comparison with the estimate provided by the central limit theorem, i.e., assuming that the left-hand side in \eqref{eq:delta_star} is equal to
\begin{equation}
\label{eq:delta_clt}
2\Phi\left(-\sqrt{M}\frac{\varepsilon}{\sigma}\right) =: \delta_{clt}
\end{equation}
where $\Phi$ is the cumulative density function of standard Gaussian distribution and $\sigma^2=\var[(\psi(g_1;\vec w)-\psi(g_2;\vec w))^2]$.
Graph $g_1,g_2$ are randomly selected from the two classes of SBM data set. Results in Figure~\ref{fig:verify-theo} show that the empirical assessments of the left-hand sides in \eqref{eq:delta_M} and  \eqref{eq:delta_star} are smaller then their respective bounds on the right-hand side of the inequalities, hence confirming the theoretical predictions.

The second experiment is conducted by comparing the performance drop in adopting the approximations~\eqref{eq:approx-distance} and \eqref{eq:approx-kernel} with varying embedding dimension $M$. The task is a binary classification, and it is performed using support vector machine and k-nearest neighbour classifiers, as standard kernel- and distance-based methods. 
Figure~\ref{fig:classif-del-sbm} shows the achieved classification accuracy. 
We see that the accuracy obtained with GRNF empirical converges to the accuracy obtained with $M_*=10^4$ features.

\subsection{GRNF as a layer of a neural net}
\label{sec:exp-gnn}

In this experiment we consider a graph network composed of GRNF as first, untrained layer of a graph neural network and we provide empirical evidence that our proposal is in line with current state-of-the-art performance in graph classification.

To this purpose, we considered the benchmark setup provided by \citet{errica2020fair} with data sets from chemical and social domains. Specifically, we considered NCI1, PROTEINS, ENZYMES, IMDB-BINARY, IMDB-MULTI and COLLAB, all available to the public \cite{KKMMN2016} and commonly used for benchmarking. The major differences between the two categories of graph are that graphs of chemical compounds come with node attributes, while the social graphs have generally higher edge density. 

Our model combines a GRNF (untrained) layer, with a linear one followed by an output layer to perform classification tasks. All intermediate layers have the rectified linear unit function $x\mapsto\max\{0, x\}$ as activation function.
We build features with $k=1,2$ tensor orders and embedding dimension $M=512$.
In Table~\ref{tab:gnn-comparison} we compare our results with those achieved by some of the most common and varied graph networks: DGCNN \cite{zhang2018end}, DiffPool \cite{ying2018hierarchical}, ECC \cite{simonovsky2017dynamic}, GIN \cite{xu2018powerful}, GraphSAGE \cite{hamilton2017inductive}). Finally, we consider a baseline%
    \footnote{We actually have two baseline models: one for PROTEINS and NCI1 which considers molecular fingerprints, and one for ENZYMES, IMDB-BINARY, IMDB-MULTI and COLLAB employing a more generic layer to aggregate node attributes.}
that has no access to the graph topology \cite{errica2020fair}.

Noteworthy, the performance of GRNF is comparable with that of the considered methods and, in most of the classification problems, is substantially better then the baseline. This shows that our proposal is in line with the current state of the art and can exploit the topological information of the graphs. 

For completeness, in Figure~\ref{fig:convergence-TU} we report the performance of GRNF on ENZYMES and IMDB-BINARY letting  the embedding dimension $M$ vary. Here, we observe that we reach a plateau in the performance with smaller embedding dimension than in the SBM and Del data set, hence fewer features were actually sufficient.

\section{Conclusions and future work}
\label{sec:conclusions}

The present paper proposes a graph embedding method that we called Graph Random Neural Features (GRNF). The method provides a way to generate expressive graph representations that preserve, with arbitrary precision, the metric structure of the original graph domain. Moreover, GRNF does not require a training phase; nonetheless, it is possible to search for a distribution $P$ that best suits the data and task at hand.
GRNF, besides providing an explicit embedding method for graphs, can be used as layer of a larger graph neural network. Finally, by approximating graph distances and kernels, GRNF can also be used in conjunction with distance- and similarity-based machine learning methods.

GRNF is based on a family $\mc F=\{\psi(\,\cdot\,;\vec w):\mc G\rightarrow \R\}$ of graph neural networks, parametrized by vector $\vec w \in\mc W$, that separates graphs of $\mc G$.
By defining a probability distribution $P$ over $\mc W$, we can sample and weight the importance of $M$ graph neural features, obtaining the proposed GRNF map $\map:\mc G\rightarrow \R^M$. 
Our results show that a distance for graphs can be obtained as the expectation of the squared discrepancy $\left(\psi(g_1;\vec w)-\psi(g_2;\vec w)\right)^2$; similarly, $\EE_{\vec w}[\xi(g_1;\vec w)\,\xi(g_2;\vec w)]$ leads to a positive-definite kernel function for graphs.
Theorem~\ref{theo:dP-metric} states that, when $\supp(P)=\mc W$ the resulting graph distance is a metric: this implies that, in principle, it is possible to distinguish between any pair of non-isomorphic graphs.
Secondly, Theorem~\ref{theo:M-to-bounds} proves that the Euclidean distance $\norm{\map(g_1)-\map(g_2)}_2$ between embedding vectors $\map(g_1),\map(g_2)$ converges to the actual graph distance $d_P(g_1,g_2)$, and provides a criterion to select an embedding dimension $M$ ensuring to preserve the metric structure of the original graph domain up to a prescribed error and probability.

We believe that investigating more sophisticated approximation methods, like Gaussian quadrature, can bring substantial improvement, especially considering the computational overhead that most of the graph processing methods require.  

\section*{Acknowledgements}
This research is funded by the Swiss National Science Foundation project 200021\_172671: ``ALPSFORT: A Learning graPh-baSed framework FOr cybeR-physical sysTems.''
The work of L. Livi was supported by the Canada Research Chairs program.


\nocite{paszke2019pytorch}

\bibliography{biblio}
\bibliographystyle{icml2020}

\clearpage

\appendix

\section{Vector space of invariant and equivariant linear graph operators}
\label{sec:def-affine-inv-equiv}

Denote with $\Gamma(k)$ the set of all partitions of $\{1,\dots, k\}$. Bell's number $\bell(k)$ represents the cardinality of $\Gamma(k)$.
Given a partition $\gamma \in \Gamma(k)$, we say that multi-index $\vec a \in \{1,\dots,n\}^k$ \emph{complies} with $\gamma$ if, for any $i,j \in \{1,\dots,n\}$, we have that
$\vec a_i = \vec a_j$ if and only if $i,j$ belong to the same set in $\gamma$.

For every $\gamma\in\Gamma(k)$, tensor $\vec I_\gamma\in\mc T^k$ is defined as $(\vec I_\gamma)_{\vec a} = 1$ for any $\vec a$ complying with $\gamma$, and $0$ otherwise. 
With the scalar tensor product
$$\vec I\cdot \vec T:=\sum_{\vec a\in\{1,\dots,n\}^k} \vec I_{\vec a} \vec T_{\vec a}\quad\in\R,$$ 
for any  $\vec I,\vec T\in\mc T^k$. We obtain that basis $\{\vec I_\gamma\}_{\gamma\in\Gamma(k)}$ is orthogonal, because the locations in which the tensors are non-null are disjoints.

Similarly, for every $\gamma\in\Gamma(k+l)$, tensor $\vec E_\gamma\in\mc T^{k+l}$ is defined as $(\vec E_\gamma)_{\vec a} = 1$ for any $\vec a$ complying with $\gamma$, and $0$ otherwise. Here the tensor product $\vec E \cdot \vec T$ between $\vec E\in\mc T^{k+l}$ and $\vec T\in\mc T^k$ is defined as 
$$\vec E\cdot \vec T:=\sum_{\vec b\in\{1,\dots,n\}^l}\sum_{\vec a\in\{1,\dots,n\}^k} \vec E_{\vec b,\vec a} \vec T_{\vec a}\quad \in\mc T^l.$$ 
Again, for different $\gamma,\gamma'$, the basis elements $\vec E_\gamma$ and $\vec E_{\gamma'}$ are orthogonal.

It is not rare that graphs come with both node and edge attributes, say $F_{node}$- and $F_{edge}$-dimensional, respectively. In this case one can create a the Cartesian product space of dimension $F=F_{node}+F_{edge}$, and represent any $n$-node graph $g$ as a tensor $\vec A_g\in\mc T^2\times \R^F$. 
Accordingly, an equivariant map is function $f:\mc T^k\times \R^F\rightarrow\mc T^l\times \R^{F'}$, so that $f(\pi\star\vec T) = \pi\star f(\vec T)$ for every $\pi\in S_n$, and where $\pi$ is now acting on all components, but the last one. Similarly, we extend the definition of invariant maps.

We refer the reader to the original paper for a detailed description~\cite{maron2018invariant}.

\section{Proofs}
\label{sec:app:proofs}

\subsection{Proof of Lemma \ref{lemma:F1-separates}}
\label{proof:lemma:F1-separates}
The proof employs the functions $f^\otimes$ in the form
\begin{multline*}
f^\otimes(g) = \sum_{s=1}^S H_{\sum k_s}\left[
    \rho_e \left( 
        F^{(s,1)}_{2,k_{s,1}}(\vec A_g;\vec \theta_{s,1})  
    \right)\right.
    \otimes\dots \\ \dots\otimes
    \rho_e \left.\left( 
        F^{(s,T)}_{2,k_{s,T}}(\vec A_g;\vec \theta_{s,1}) 
    \right);\vec \theta_H
\right] + b
\end{multline*}
with functions $\{H_{i}^{(s)}\}$ and $\{F_{2,j}^{(s,t)}\}$ linear invariant and equivariant functions as defined in \eqref{eq:affine-inv-equiv}, but without the bias terms. We denote with $\mc N^\otimes(\rho_e)$ the set of such functions letting $S,T$ vary in $\mathbb{N}$, $b\in\R$, and for any $(k_{1,1},\vec \theta_{s,t},\vec \theta_H)\in\mc W$. We also denote with $\mc N(\rho_e)$ the restriction of $\mc N^\otimes(\rho_e)$ to $T=1$, which is contained in the closure under finite sums of set $\mc F(\id, \rho_e)$, where $\id:\R\rightarrow \R$ is the identity function.

\begin{enumproof}
\item 
\citet{keriven2019universal} showed that $\mc N^\otimes(\sigma)$, with $\sigma$ the sigmoid activation, separates $\mc G$ [Lem. 2] and that $\mc N(\rho_e)$ is dense in $\mc N^\otimes(\sigma)$ for any squashing function $\rho_e$ [Lem. 3].
Since $\mc N(\rho_e)\subseteq \mc F({\id},\rho_e)$, then $\mc F({\id},\rho_e, \mc W)$ is dense on $\mc N^\otimes(\rho_e)$, and it derives that $\mc F({\rm id},\rho_e, \mc W)$ separates points of $\mc G$, as well.
We conclude that for any pair of distinct graphs $g_1\ne g_2$ in $\mc G$, there is a function $\tilde f \in\mc F(\id,\rho_e, \mc W)$, such that $\tilde f(g_1)\ne \tilde f(g_2)$.

\item 
Notice that, for any $a,b\in\R$, $f_{a,b}(\cdot):=a \tilde f(\cdot)+b\in\mc F(\id,\rho_e, \mc W)$, and when $a\ne 0$ we also have $f_{a,b}(g_1)\ne f_{a,b}(g_2)$; therefore, we can push $f_{a,b}(g_1)$ and $f_{a,b}(g_2)$ to any
desired location. Let $\delta:=\tilde f(g_2)-\tilde f(g_1)>0$ and, without loss on generality, assume that $\rho_i(0)\ne \rho_i(1)$ (in fact, $\rho_i$ is non-constant). With the choice $a=\frac{1}{\delta}$ and $b=-\frac{\tilde f(g_2)}{\delta}$, we obtain that $f_{a,b}(g_2)=0$ and $f_{a,b}(g_1)=1$ and $\rho_i(f_{a,b}(g_2))\ne\rho_i(f_{a,b}(g_1))$, which proves that for any $g_1\ne g_2$ there exists a function $f\in\mc F(\rho_i,\rho_e,\mc W)$ such that $f(g_1)\ne f(g_2)$.
\end{enumproof}

\subsection{Proof of Theorem~\ref{theo:dP-metric}}
\label{proof:theo:dP-metric}

\begin{enumproof}
\item 
Since the $\delta(g_1,g_2):=(\psi(g_1)-\psi(g_2))^2\ge 0$, $\delta(g_1,g_2)=\delta(g_2,g_1)$ and $\delta(g_1,g_1)=0$ for any $g_1,g_2\in\mc G$, then the same properties hold also for $d_P(g_1,g_2)=\sqrt{\EE[\delta(g_1,g_2)]}$.

\item
The Cauchy-Schwarz inequality
\begin{align}
\label{eq:CS-inequality}
|\EE[X_1X_2]|^2\leq \EE[X_1^2]\EE[X_2^2],
\end{align} 
holds for any pair of random variables $X_1,X_2$; in fact, notice that 
\begin{multline*}
0\leq \frac{1}{2}\,\EE\left[\left(\tfrac{X}{\sqrt{\EE[X^2]}}-\tfrac{Y}{\sqrt{\EE[Y^2]}}\right)^2\right]
\\= 1 -  \frac{\EE\left[X Y\right]}{\sqrt{\EE[X^2]}\sqrt{\EE[Y^2]}};
\end{multline*}

\item 
By \eqref{eq:CS-inequality}, we have
\begin{multline}
\label{eq:triangular-expected}
\EE[(X_1+X_2)^2]\leq \EE[X_1^2] +2 \sqrt{\EE[X_1^2]\EE[X_2^2]} +\EE[X_2^2]
\\=\left(\sqrt{\EE[X_1^2]}+\sqrt{\EE[X_2^2]}\right)^2.
\end{multline}
The triangular inequality follows from the choice $X_1=\psi(g_1;\vec w)-\psi(g_3;\vec w)$ and $X_2=\psi(g_3;\vec w)-\psi(g_2;\vec w)$.

\item 
Finally, the identifiability property \eqref{eq:identifiability} is proved by the following Lemma~\ref{lemma:dP-identifiability}.
\end{enumproof}

\begin{lemma}
\label{lemma:dP-identifiability}
Under Assumptions \ref{ass:rho} and \ref{ass:suppP}, we have that for any pair of graphs $g_1,g_2\in\mc G$
$$ g_1= g_2 \iff d_{P}(g_1, g_2)=0.$$
\end{lemma}
\begin{proof}
Denote with $\mc W_k$ the parameter set $\mc W$ in which the hidden tensor order is fixed to $k$.
\begin{enumproof}

\item Assumption~\ref{ass:rho} enables Lemma \ref{lemma:F1-separates}, therefore for any pair $g_1,g_2\in\mc G$ with $g_1\ne g_2$ there exists a parameter configuration $\tilde{\vec w}\in\mc W$ so that $\psi(g_1,\tilde{\vec w})\ne \psi(g_2,\tilde{\vec w})$.

\item Again from Assumption~\ref{ass:rho}, for any $k\in\mathbb{N}$, we have that $\psi(g;\cdot):\mc W\rightarrow \R$ limited to $\mc W_k$, is continuous, as it is the composition of linear operators and continuous activations $\rho_i,\rho_e$. This holds in particular for $\tilde k$, the hidden tensor order associated with $\tilde{\vec w}$. Therefore, there is a neighbourhood $U_{g_1,g_2}(\tilde{\vec w})$ of $\tilde{\vec w}$ such that 
$$
|\psi(g_1,\vec w)-\psi(g_2,\vec w)| \geq \frac{\varepsilon_{g_1,g_2}}{2},\qquad \forall\vec w\in U( \tilde{\vec w}),
$$
with $\varepsilon_{g_1,g_2}=|\psi(g_1;\tilde{\vec w})-\psi(g_2;\tilde{\vec w})|>0$.

\item 
Assumption~\ref{ass:suppP} ensures that $\supp(P)=\mc W$, and that $P(U_{g_1,g_2}(\tilde{\vec w}))$ is strictly positive, independently on the choice of graphs $g_1,g_2$.
We conclude that for any pair $g_1,g_2$ of distinct graphs, $$d_P(g_1,g_2)\geq P(U_{g_1,g_2}(\tilde{\vec w})) \frac{\varepsilon_{g_1,g_2}}{2} > 0.$$
\end{enumproof}
\end{proof}

\section{Limiting the order of the hidden tensor: Weighted GRNF}
\label{sec:wGRNF}

Allowing order $k$ to grow indefinitely might result in an infeasible computation load.
In the following section, we show how to cope with this problem by defining a $d_P$ and $\kappa_P$ over a distribution $P$, while sampling parameter $\vec w$ from a different and more convenient one, $\overline P$.

Limiting $k$ to be less or equal than $k^*$ results in distance $d_{P}(\cdot,\cdot)$ which is not metric, in general, and one can build practical counterexamples. 
Consider a $P$ with $\supp(P)=\mc W$, and assume $p$ to be the marginal probability mass function associated to $k$. Consider also a nonempty subset $\mc K\subseteq\mathbb{N}$ and define the probability function $\overline P$ with marginal probability mass function
$$\bar{p}(k) = \frac{p(k)}{P(\mc K)},\qquad k\in\mc K, \qquad \text{ and $0$ otherwise}$$
where $P(\mc K)$ is the normalizing factor $\sum_{l\in\mc K}p(l)$.
We obtain that approximating $\kappa_P$ and $d_P^2$ by sampling the parameters $\vec w$ from $\overline P$, and considering the following modified GRNF
\begin{equation}
\label{eq:GRNF-bounded-order}
\overline\map(\;\cdot\;;\vec W)
:=\sqrt{P(\mc K)}\;\map(\;\cdot\;;\vec W),
\end{equation}
yields a practical alternative, as shown in the following lemma. We call $\overline \map$ in \eqref{eq:GRNF-bounded-order} \emph{bounded-order GRNF} to distinguish it from the \emph{plain} GRNF \eqref{eq:GRNF-plain}.
\begin{lemma}
\label{lemma:expectation-reduced-p}
Consider the bounded-order GRNF \eqref{eq:GRNF-bounded-order}. If $\rho_i$ is bounded by a constant $C_{\rho_i}$, then
\begin{multline*}
\EE\left[\norm{\overline \map(g_1;\vec W)-\overline \map(g_2;\vec W)}_2^2\right]
\\\begin{cases}
\geq d_{P}(g_1, g_2)^2 - (1-\prob_p(\mc K))\,4C_{\rho_i}^2 
\\
\leq d_{P}(g_1, g_2)^2.
\end{cases}
\end{multline*}
\end{lemma}
\begin{proof}
\begin{enumproof}

\item Let us start with a generic random variable $X$.
\begin{align*}
\EE_{\overline p}\left[X\right] 
  &=\sum_{k=1}^\infty \overline p(k) X 
   =\sum_{k\in\mc K} \overline p(k) X 
   =\frac{1}{\prob_p(\mc K)}\sum_{k\in\mc K} p(k) X \\
  &=\frac{1}{\prob_p(\mc K)}  \sum_{k=1}^\infty p(k) X - \sum_{k\not\in\mc K} p(k) X \\
  &=\frac{1}{\prob_p(\mc K)} \EE_p[X] - \frac{1}{\prob_p(\mc K)}\sum_{k\not\in\mc K} p(k) X.
\end{align*}

\item
Substituting $X=(\map(g_1;\vec w)-\map(g_2;\vec w))^2$, 
and then taking the expectation with respect to the joint $\overline P$, we get
$\EE_{\overline P}[X]\le d_P(g_1,g_2)^2$. 
Finally, being $\rho_i$ bounded by a constant $C_{\rho_i}$, we get $X\leq 4C_{\rho_i}^2$, hence the thesis.
\end{enumproof}
\end{proof}

We stress that, despite the hidden orders are sampled from the bounded-order distribution $\overline p$, the result relates to distance $d_{P}(\cdot,\cdot)$, which is with respect to the original distribution $P$ with $\supp(P)=\mc W$.

As we can see, completely avoiding certain hidden tensor orders $k$ comes with the price of biased estimations, which does not ensure convergence in \eqref{eq:approx-kernel} and \eqref{eq:approx-distance}. One can obtain consistent approximations \eqref{eq:approx-kernel} and \eqref{eq:approx-distance} while mitigating the computational and memory burden by selecting probability distribution $\overline P$, which down-weights large hidden-tensor orders maintaining $\supp(\overline P)=\mc W$. 
We can also make a step further and let the entire distribution $P$ vary. We result in the Weighted GRNF defined in \eqref{eq:GRNF-weighted}. This type of embedding is a generalization of both the plain GRNF \eqref{eq:GRNF-plain} and the bounded-order GRNF \eqref{eq:GRNF-bounded-order}.

We prove that a generalized version of Theorem~\ref{theo:M-to-bounds} that applies to the weighted GRNF. 
\begin{theorem}
\label{theo:M-to-bounds-generalized}
Consider a distribution $\overline P$ over $\mc W$, with $\supp(\overline P)=\mc W$. If there exists a positive constant $\overline \CEfour$ such that the fourth momentum
$$\EE_{\vec w\sim \overline P}\left[\frac{p(\vec w)^2}{\overline p(\vec w)^2}\psi(g;\vec w)^4\right]<\overline \CEfour$$
for any choice of $g\in\mc G$, 
then for any value $\varepsilon>0$ and $\delta\in(0,1)$, 
when $M\geq \frac{16\overline\CEfour}{\delta\,\varepsilon^2}$ we have
$$\prob\left(\norm{\norm{\overline\map(g_1)-\overline\map(g_2)}_{2}^2 -d_{P}(g_1,g_2)^2} \ge \varepsilon\right) \le \delta.$$
\end{theorem}
\begin{proof}
Following the rationale of the proof of Theorem \ref{theo:M-to-bounds}, we prove that $\EE[\norm{\overline\map(g_1)-\overline\map(g_2)}_{2}^2]=d_{P}(g_1,g_2)^2$ and that $\EE[\norm{\overline\map(g_1)-\overline\map(g_2)}_{2}^2]$ scales as $O(M^{-1})$. Finally, we apply the Chebyshev's inequality. 
\begin{enumproof}
\item
Denote with $\Delta(\vec w)=\norm{\psi(g_1;\vec w)-\psi(g_2;\vec w)}_{2}^2$.
\begin{align*}
&\EE_{\overline P^M}\left[\norm{\overline\map(g_1)-\overline\map(g_2)}_{2}^2\right] = 
\\&\qquad\qquad=\sum_{m=1}^M \EE_{\overline P}\left[\frac{1}{M}\frac{p(\vec w_m)}{\overline p(\vec w_m)}\Delta(\vec w_m)\right]
\\&\qquad\qquad=\frac{1}{M}\sum_{m=1}^M \int_{\mc W}\frac{p(\vec w)}{\overline p(\vec w)}\;\Delta(\vec w)\;{\rm d}P(\vec w)
\\&\qquad\qquad=\int_{\mc W}\overline p(\vec w)\;\frac{p(\vec w)}{\overline p(\vec w)}\;\Delta(\vec w) \;{\rm d}\vec w
\\&\qquad\qquad=\EE_P\left[\Delta(\vec w)\right]=d_P(g_1,g_2)^2
\end{align*}
This holds thanks to the fact that $\overline p(\vec w)\ne 0$ for every $\vec w\in\mc W$, otherwise we would end up with a result similar to Lemma~\ref{lemma:expectation-reduced-p}.

\item
The variance can be bound in the same manner of \eqref{eq:bound-variance}, obtaining
\begin{multline*}
\var\left[\norm{\overline\map(g_1)-\overline\map(g_2)}_2^2\right] 
  \\= \frac{1}{M} \var\left[\frac{p(\vec w)}{\overline p(\vec w)}\left(\psi(g_1;\vec w)-\psi(g_2;\vec w)\right)^2\right]
\\\leq\frac{16\,\overline\CEfour}{M}
\end{multline*}

\item 
Chebyshev's inequality gives us the bound
\begin{align*}
\prob_{\overline P^M}\left(\norm{\norm{\overline\map(g_1)-\overline\map(g_2)}_2^2-d_P(g_1,g_2)}\geq \varepsilon \right) \leq 
\frac{16\,\overline\CEfour}{M\,\varepsilon^2}
\end{align*}
from which the thesis follows.
\end{enumproof}
\end{proof}

\section{Computational complexity}
\label{sec:comp-complexity}

Let us consider $M$ random features with hidden tensor of order $k$.
The computational complexity of \eqref{eq:GRNF-plain} is given by:
\begin{itemize}
    \item 
    $\bell(2+k)$ operations of the form $\vec E_\gamma \vec T$, each with cost $O(n^{2+k} \, F\, F_h)$;

    \item 
    then, we perform $M$ linear combinations $\sum_\gamma \theta_\gamma \vec I_\gamma T$ and addition of the bias term, each with cost $ O((2 \bell(2+k)+1) n^{k} \, F_h)$;
    
    \item
    in order to compute $H_k(\vec T)$ \eqref{eq:affine-inv-equiv} for each of the $M$ features, we perform $\bell(k)$ operations of the form $\vec I_\gamma T$, which scale as $O(n^k \cdot F_h)$. Considering also the linear combination $\sum_\gamma \theta_\gamma \vec I_\gamma T$ and the bias term $\theta'$, we have $O(\bell(k)\, (n^{k} \, F_h + 1) + 1)$.

\end{itemize}
The total computational complexity for creating a graph representation is:
\begin{multline*}
O(\bell(2+k)n^{2+k} \, F\, F_h)
\\+
O(M (2 \bell(2+k)+1) n^{k} \, F_h)
\\+
O(M \bell(k)\, (n^{k} \, F_h + 1) + M).
\end{multline*}
which is equivalent to 
\begin{multline*}
\label{eq:final-comp-complexity}
O\big(\bell(2+k)\,n^{2+k} \, F\, F_h
\\ +M\,n^{k}\, F_h\,( \bell(2+k)+ \bell(k) )\big).
\end{multline*}

\section{Implementation details}

\paragraph*{GRNF implementation}

A PyTorch \cite{paszke2019pytorch} implementation of GRNF is available at the following link \url{https://github.com/dzambon/graph-random-neural-features} and adopts the efficient version for $k=1,2$ described in \cite{maron2018invariant}. When not specified, $\rho_e(x)=\max\{0, x\}$ is the rectified linear unit, $\rho_i(x)={\rm tanh}(x)$ is the hyperbolic tangent, $F_h=4$ features in the hidden tensor, the probability of having order $k=1$ and $k=2$ in the hidden tensor is $2/3$ and $1/3$, respectively, and the weights $\theta_F,\theta_H$ drawn from a standard Gaussian distribution. 
The provided implementation can run on ordinary laptops.

\paragraph*{Replicability of the experiments}
The source code for running all the synthetic experiments is available at the GRNF repository. All the other experiments are performed with the framework provided by \citet{errica2020fair} at the repository \url{https://github.com/diningphil/gnn-comparison}.
All competitor models considered in our study are set up with the hyper-parameters suggested in \cite{errica2020fair}.

\end{document}